\documentclass[letterpaper, 10 pt, conference]{ieeeconf}  

\IEEEoverridecommandlockouts                              

\usepackage{xcolor}
\usepackage[table]{xcolor}
\usepackage{booktabs}
\usepackage{bm}
\usepackage{multirow}
\usepackage{multicol}
\usepackage{nicematrix}
\usepackage{threeparttable}
\usepackage{amssymb}
\usepackage{hyperref}
\usepackage{cleveref}
\let\labelindent\relax
\usepackage{caption}
\DeclareCaptionLabelSeparator{dottilde}{.~~}
\usepackage{enumitem}
\usepackage{subfigure}
\usepackage{algorithmic}
\usepackage{cuted}
\usepackage{capt-of}

\definecolor{myemerald}{rgb}{0.753, 0.898, 0.804}
\definecolor{mylightgreen}{rgb}{0.894, 0.933, 0.745}
\definecolor{myyellow}{rgb}{0.996, 0.972, 0.780}

\newcommand{\rt}[1]{\textcolor{red}{#1}}
\newcommand{\bt}[1]{\textbf{#1}}
\newcommand{\ti}[1]{\textit{#1}}

\newcommand{\firstc}{\cellcolor{myemerald!100}}
\newcommand{\secondc}{\cellcolor{mylightgreen!100}}
\newcommand{\thirdc}{\cellcolor{myyellow!100}}
\newcommand{\firsttxt}[1]{\colorbox{myemerald}{#1}}
\newcommand{\secondtxt}[1]{\colorbox{mylightgreen}{#1}}
\newcommand{\thirdtxt}[1]{\colorbox{myyellow}{#1}}

\author{
Jiamin Zheng$^*$, Jingwen Yu$^*$, Guangcheng Chen, and Hong Zhang$^{\dagger}$
\thanks{All authors are with the Shenzhen Key Laboratory of Robotics and Computer Vision, Southern University of Science and Technology, Shenzhen, China.}
\thanks{Jingwen Yu is also with the CKS Robotics Institute, Hong Kong University of Science and Technology, Hong Kong SAR, China.}
\thanks{$^{\dagger}$ corresponding author (hzhang@sustech.edu.cn), $^*$ equal contribution.}
\thanks{This work was supported in part by Shenzhen Science and Technology Program (No. SGDX20240115111759002), in part by Meituan Academy of Robotics Shenzhen, in part by the Shenzhen Association for Science and Technology (No. XHXS2025-003), and in part by High level of special funds (G03034K003) from Southern University of Science and Technology, Shenzhen, China.}
}

\begin{document}
\title{\LARGE \bf
Enhancing Glass Surface Reconstruction via Depth Prior \\ for Robot Navigation}

\makeatletter
\let\@oldmaketitle\@maketitle
\renewcommand{\@maketitle}{\@oldmaketitle
\centering
\captionsetup{type=figure, singlelinecheck=false,font=small}
\begin{tabular}{cccc}
\includegraphics[width=0.95\textwidth,trim=5mm 17mm 10mm 0mm]{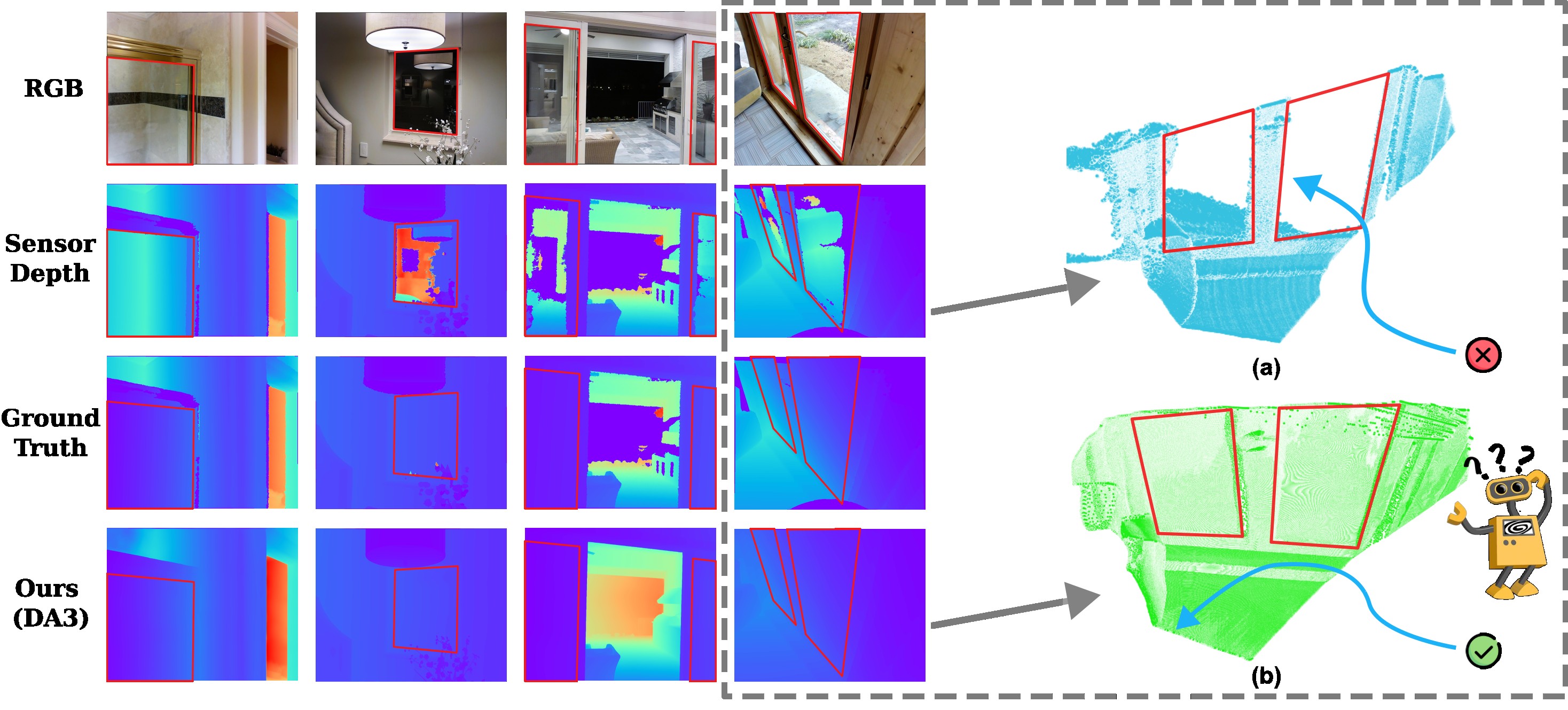}
\end{tabular}
\caption{Our method, combined with Depth Anything 3 (DA3)~\cite{depthanything3}, achieves superior performance on glass surface (\rt{red bounding box}) depth estimation compared to the raw depth sensor measurements, providing accurate geometry for robot navigation in indoor environments. A 3D point cloud visualization of the scene in the last column is presented. (a) points reconstructed from the sensor depth, and (b) points from our estimated depth. In (a), the robot might misperceive the glass region as traversable, which leads to a collision. 
}
\label{fig:teaser}
\setcounter{figure}{1}
}

\makeatother
\maketitle
\begin{abstract}
Indoor robot navigation is often compromised by glass surfaces, which severely corrupt depth sensor measurements. While foundation models like Depth Anything 3 provide excellent geometric priors, they lack an absolute metric scale. We propose a training-free framework that leverages depth foundation models as a structural prior, employing a robust local RANSAC-based alignment to fuse it with raw sensor depth. This naturally avoids contamination from erroneous glass measurements and recovers an accurate metric scale. Furthermore, we introduce \ti{GlassRecon}, a novel RGB-D dataset with geometrically derived ground truth for glass regions. Extensive experiments demonstrate that our approach consistently outperforms state-of-the-art baselines, especially under severe sensor depth corruption. The dataset and related code are available at https://github.com/JMamie/GlassRecon.
\end{abstract}
\section{Introduction}
Reliable geometric perception is fundamental to localization, mapping, and safe navigation across diverse robotic platforms and environments~\cite{wei2025fusionportablev2}. 
Glass surfaces represent a particularly hazardous failure case for standard RGB-D sensors, which often produce invalid measurements or incorrectly capture background objects.
Existing solutions, such as specialized LiDAR, complementary sensors, or glass-specific neural networks, are often constrained by environmental conditions, high hardware costs, or poor generalizability to unseen domains.

While modern monocular depth estimation models (e.g., Depth Anything V3~\cite{depthanything3}) provide powerful structural priors, they fail to deliver accurate metric scale on their own. Besides the lack of absolute scale in affine-invariant networks, state-of-the-art (SotA) metric-depth networks still struggle to accurately estimate the geometry and scale of glass surfaces.

To bridge this gap, we propose a modular, training-free pipeline that leverages a modern affine-invariant monocular network to obtain a structural depth prior. We then align this prior to the sensor's metric scale using a novel local RANSAC-based alignment. By calculating scale-shift pairs after local sampling from image patches and validating them globally, our method inherently avoids contamination from erroneous sensor measurements on glass, preserving the prior's structural fidelity.

To rigorously evaluate our approach, we introduce \ti{\bt{GlassRecon}}, a dedicated dataset of $\sim1K$ indoor RGB-D images featuring glass instances. Assuming most indoor glass is planar, we generate ground-truth depth using geometric constraints derived from reliable coplanar surfaces. The dataset features an ``\ti{easy} / \ti{hard}'' split based on the severity of sensor depth corruption, enabling nuanced evaluation.

Our main contributions are summarized as follows:
\begin{itemize}
    \item \textbf{Modular Pipeline:} A robust, training-free depth completion method that combines monocular depth priors with local RANSAC alignment.
    \item \textbf{Benchmark Dataset:} The \ti{\bt{GlassRecon}} dataset ($\sim1K$ annotated images) with geometrically derived ground truth and \ti{easy} / \ti{hard} splits for standardized benchmarking.
    \item \textbf{SotA Performance:} Extensive experiments demonstrating that our method consistently outperforms global alignment baselines and metric depth prediction networks, with particularly significant gains on \ti{hard} examples.
\end{itemize}
\section{Related Work}
\label{sec:related}

\subsection{Glass Detection for Robot Navigation}
Glass surfaces present a critical challenge for robot navigation, often causing conventional depth sensors to produce erroneous or missing measurements. Various approaches address this issue across different modalities.

\textbf{LiDAR-based methods.} LiDAR can detect glass by exploiting specular reflections at specific incidence angles \cite{wang2017detecting, zhou2024lidar}. However, their reliability heavily depends on favorable viewing conditions (e.g., near-perpendicular incidence and close range), which limits their general applicability in unconstrained robotic navigation.

\textbf{Sensor fusion-based methods.} Multi-modal approaches combine depth cameras with complementary sensors such as ultrasonic \cite{zhang2017three}, polarization \cite{mei2022glass}, or thermal \cite{huo2023glass} cameras. While these modalities improve detection robustness regardless of optical transparency, they introduce hardware complexity, increase system cost, and often lack the density or speed required for real-time navigation.

\textbf{Data-driven methods.} Recent SLAM systems integrate deep learning-based glass segmentation \cite{zhu2021transfusion, zhao2023glass} to exclude unreliable points during pose estimation, subsequently reconstructing the glass separately. Concurrently, datasets like RGB-D GSD \cite{lin2025leveraging}, GW-Depth \cite{liang2023monocular}, and MonoGlass3D \cite{zhang2025monoglass3d} have been introduced to train glass-specific segmentation and depth networks. 

Unlike these approaches, which are inherently tied to dedicated training data and specific glass types, our method requires neither glass-specific training data nor specialized hardware. By robustly aligning general-purpose monocular priors, our approach is highly generalizable and readily deployable.

\subsection{Monocular Depth Estimation}
Monocular depth estimation provides rich geometric priors for surface reconstruction. We review two relevant categories of these methods.

\textbf{Affine-invariant depth estimation.} Models such as MiDaS \cite{ranftl2020towards} and the Depth Anything series \cite{yang2024depth-1, yang2024depth, depthanything3} predict robust relative depth (or disparity) up to an unknown scale and shift. They capture detailed scene structures and generalize exceptionally well to unseen domains. However, because they lack metric information, they cannot directly provide the absolute depth required for navigation. Our method harnesses their structural accuracy by aligning them to metric sensor depth using only reliable regions.

\textbf{Metric depth estimation.} Recent networks aim to directly recover absolute scale from single images. Methods like Metric3Dv2 \cite{hu2024metric3d}, UniDepthV2 \cite{piccinelli2025unidepthv2}, and MoGe-2 \cite{wang2025moge2} tackle varying camera intrinsics via canonical spaces or self-promptable modules to output metric depth. Furthermore, depth completion networks like PriorDA \cite{wang2025depth} refine metric priors using RGB guidance. Despite these advances, metric networks frequently fail on glass due to incorrect scale estimation or missing geometry, and completion networks remain heavily constrained by corrupted input priors. 

Our approach bridges this gap by aligning highly accurate affine-invariant structural priors to reliable sensor measurements. This effectively combines the structural fidelity of modern monocular networks with the metric accuracy of RGB-D sensors, successfully recovering glass surfaces without requiring network modifications.
\begin{figure}[!t]
\centering
\includegraphics[width=0.48\textwidth]{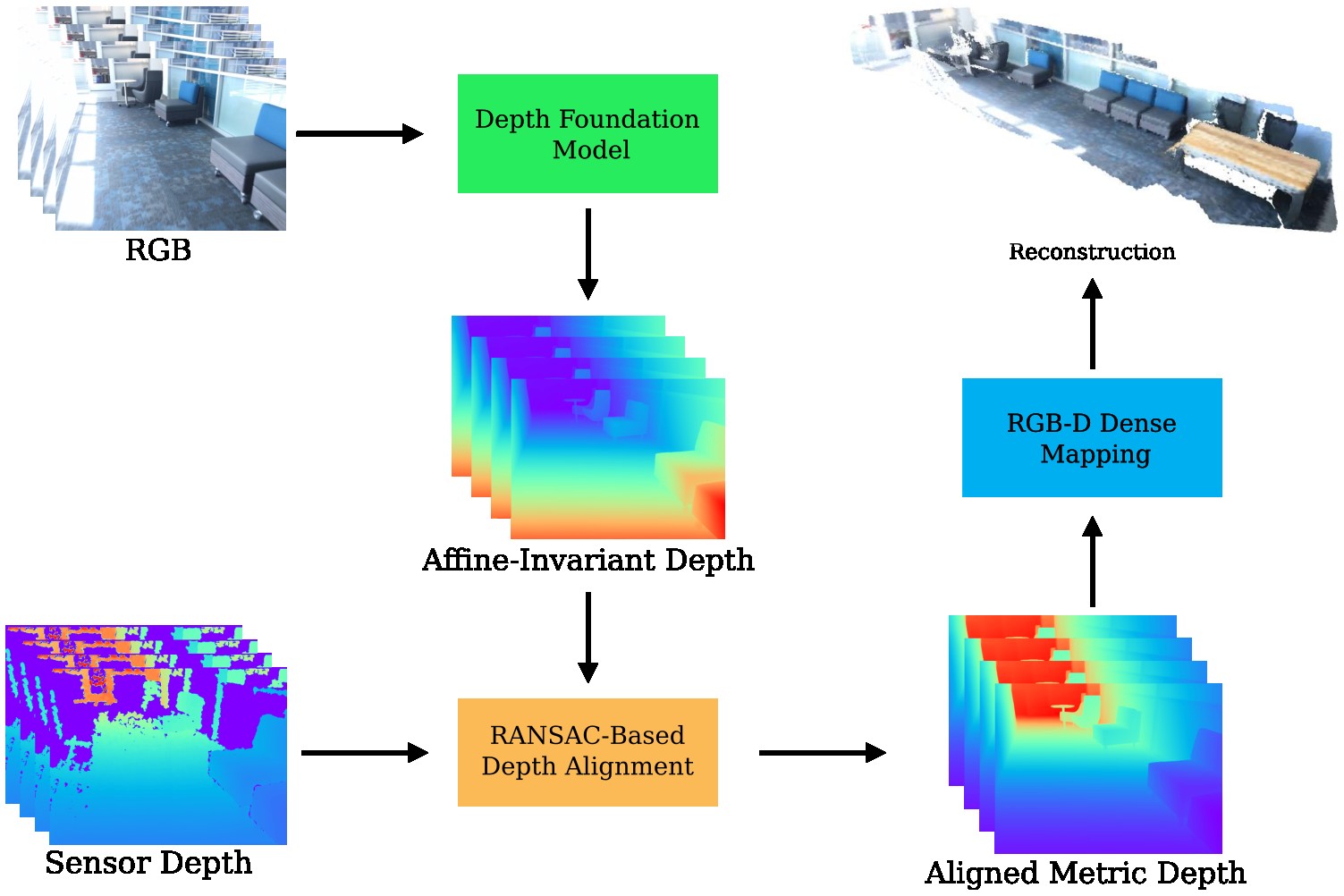}
\caption{\textbf{System pipeline.} From RGB-D inputs, a foundation model (e.g., DA3~\cite{depthanything3}) predicts affine-invariant depth. Our \ti{\bt{RANSAC-based depth alignment}} (Sec.~\ref{sec:ransac}) fuses this with raw sensor measurements to estimate the metric scale, yielding accurate dense mapping for robot navigation.}
\label{fig:pipeline}
\end{figure}

\section{GlassRecon}
\label{sec:anno}
\begin{figure}[!ht]
\centering
\includegraphics[width=0.48\textwidth]{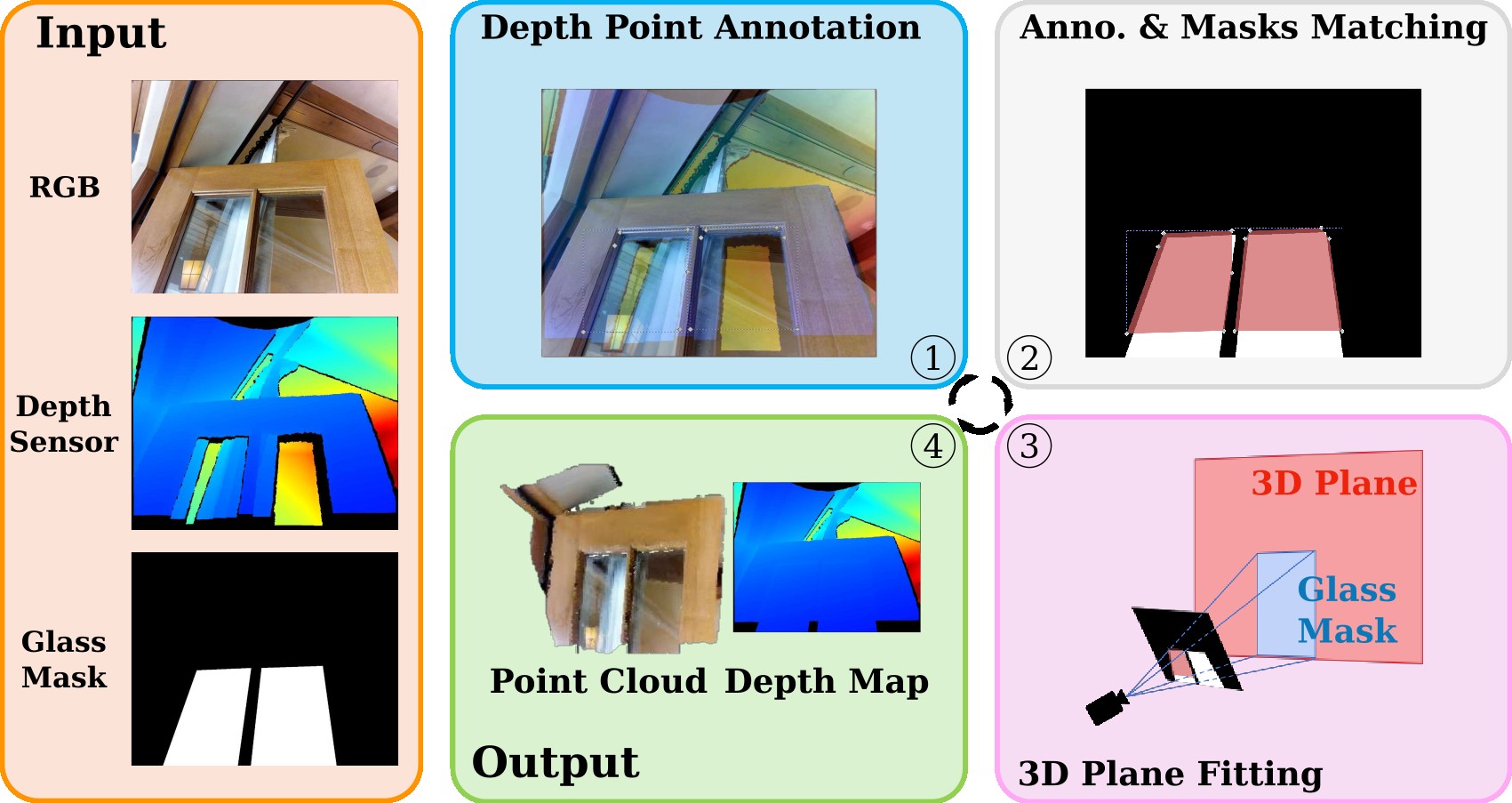}
\caption{\textbf{Annotation pipeline.} As discussed in Sec.~\ref {sec:anno}, the complete annotation process takes original RGB, depth sensor measurements, and glass mask as input, and produces a metrically aligned depth map.}
\label{fig:anno}
\end{figure}

To evaluate the proposed method and support future research on glass surface reconstruction, 
we proposed a dataset \ti{\bt{GlassRecon}} of indoor scenes that contains glass surface structures with corresponding accurate annotated depth maps. The datasets are curated from existing RGB-D datasets with further detailed annotations as shown in Fig.~\ref{fig:anno}. Our proposed dataset serves as a comprehensive benchmark for glass surface depth estimation. 

\subsection{Data Sources}
We assume that most glass structures in indoor environments (e.g., windows, glass doors, and tabletops) are planar surfaces. We source images from two publicly available datasets: Matterport3D~\cite{Matterport3D}, which provides RGB-D data and camera intrinsics, and RGB-D GSD~\cite{lin2025leveraging}, which contains images with manually labeled glass surfaces selected from Matterport3D. Based on RGB-D and glass masks, we construct a dedicated RGB-D Glass Surface Dataset through a multi-step processing pipeline that generates complete depth maps for glass regions. As shown in Fig.~\ref{fig:teaser}, the raw sensor depth map fails to accurately measure the depth of glass surfaces, so further annotation is required. 

\subsection{Annotation Pipeline}
As shown in Fig.~\ref{fig:anno}, the complete annotation contains four steps: i) depth point annotation, ii) annotation and glass masks matching, iii) 3D plane fitting, and iv) depth estimation via ray-plane intersection.

\subsubsection{Depth Point Annotation}
For each glass instance, we annotate depth points that lie on the same physical plane as the glass. Three or more pixels on reliable regions coplanar with the glass (e.g., for a window, we select points on the window frame) are selected, which can be used to define the true depth of the glass surface.

\subsubsection{Annotations and Glass Masks Matching}
For each sample, RGB-D GSD provides a binary mask indicating individual glass regions. Since multiple glass instances may exist for a single image, the correspondences between each set of annotated depth points and masks must be established. 
To achieve this, we compute the convex hull of the annotated points and measure its overlap with each mask. The mask that yields the highest intersection-over-hull ratio is considered to correspond to the glass instance. This ensures that the annotated points are spatially consistent with the corresponding glass region and represent the underlying glass plane.
\begin{figure}[!h!t]
\centering
\includegraphics[width=0.45\textwidth]{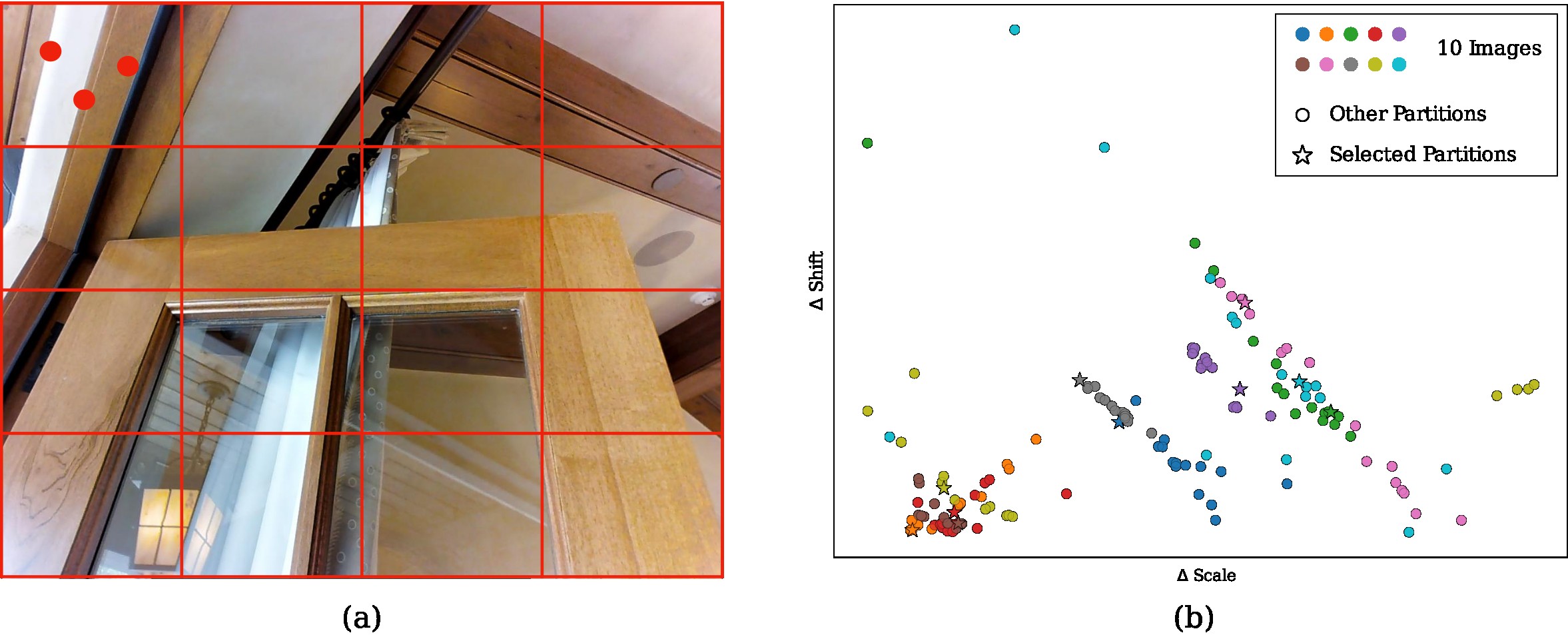}
\caption{\textbf{Patch-wise sampling strategy.} 
(a) The image and its depth map are divided into grid patches $\{\mathbf{p}_{ij}\}$ and within each patch, the $M$ pixels (\rt{red points}) are randomly selected to compute $s_{ij}$ and $t_{ij}$. (b) 10 images are randomly sampled from the proposed dataset and serve as representatives. The distributions of the estimated $s$ and $t$ over all partitions form distinct intra-image clusters (circles). The best selected (star) partition successfully anchors the geometric center of each image's distribution.} 
\label{fig:method}
\end{figure}

\subsubsection{3D Plane Fitting}
The sampled 2D image coordinates are lifted into 3D space via inverse perspective projection using the camera intrinsic matrix. This yields a set of spatial coordinates $\mathbf{P}_i = (x_i, y_i, z_i)^T$ representing the glass boundary. We model this geometry by minimizing the algebraic distance of the points to a parameterized plane:
\begin{equation}
    \label{equ:plane}
    \mathbf{n}^T \mathbf{P}_i + d = 0
\end{equation}
where $\mathbf{n} = (a, b, c)^T$ denotes the surface normal vector. The optimal plane parameters are derived under the strict unit-norm constraint $||\mathbf{n}||_2 = 1$

\subsubsection{Depth Estimation via Ray-Plane Intersection}
For every pixel $\{u,v\}$ inside the glass mask, we compute its corresponding ray direction $\mathbf{v_{uv}} = K_{c}^{-1} [u,v,1]^T$ given camera intrinsics. The depth value $z_{uv}$ at that pixel is then obtained by intersecting the ray with the fitted plane:
\begin{equation}
\label{equ:ray}
    \mathbf{P_{uv}} = \mathbf{o} + \lambda_{uv} \mathbf{v_{uv}}, \quad
    \text{s. t.} \quad \mathbf{n}^T \mathbf{P}_{uv} + d = 0
\end{equation}
where $\mathbf{o}$ denotes the camera's optical center, which we assume to be at the coordinate origin without loss of generality. Solving the system for the $\lambda_{uv}$ directly yields the metric depth along the ray.

\subsubsection{Final Depth Image Generation}
The computed depths replace the original depth values within the glass mask, producing a corrected depth image. If a glass region lacks any depth point annotations (e.g., due to severe sensor failure), that area is masked and excluded from evaluation.

To ensure the reliability of the ground truth, each image’s corrected depth is converted to a 3D point cloud and manually inspected. Only images where the glass plane appears geometrically consistent are retained. In total, our dataset comprises 917 images with glass surfaces, each accompanied by: a) RGB image, b) raw sensor depth map, c) binary glass mask, d) annotated depth (ground truth used for evaluation in Sec.~\ref{sec:exp} and e) camera intrinsics.)

\ti{\bt{GlassRecon}} provides a challenging benchmark for glass surface depth completion and reconstruction, with downstream applications in robot navigation and scene understanding.

\section{Methodology}
\subsection{Overview}
We propose a lightweight and modular pipeline for correcting depth measurements for glass surfaces, as illustrated in Fig. \ref{fig:pipeline}. The core idea is to leverage the dense, structure-aware predictions of monocular depth foundation models (DFMs) and align them to the absolute metric scale of the depth sensor using a robust patch‑wise RANSAC approach as detailed in Sec.~\ref{sec:ransac}.
Let $I$ be an RGB image and $D_{raw}$ be the corresponding raw depth map from an RGB‑D sensor (e.g., \ti{RealSense}, \ti{Kinect}). Our pipeline consists of the following three stages:
\subsubsection{Monocular Depth Prior} Any affine‑invariant monocular depth estimation network $\mathcal{F}$ can be employed to predict a relative depth map $D_{prior} = \mathcal{F} (I)$. This prior captures detailed scene structure but lacks metric scale and shift. Examples include DAV2~\cite{yang2024depth}, DA3~\cite{depthanything3}, MoGe~\cite{wang2025moge}. Through exhaustive experiments on \ti{\bt{GlassRecon}}, we found that DFMs that are trained with extremely large data learn the glass surface depth accurately while preserving generalizability on other material surfaces, specifically, DA3, as shown in Table~\ref{tab:exp1} demonstrates superior performance among SotA monocular DFMs.
\subsubsection{Affine Alignment} We estimate a global scale $s$ and shift $t$ that map the relative prior $D_{prior}$ to the raw sensor depth $D_{raw}$. 
The transformation $D_{aligned} = sD_{prior} + t$ is computed using a RANSAC‑based procedure that samples candidate parameters from local image patches and selects the pair that minimizes the alignment error across the entire image. 
This step is robust to the presence of erroneous glass measurements, as shown in Fig.~\ref{fig:method}~(b). The majority of the reliable patches outvote the minority of erroneous regions.

\subsubsection{Downstream Reconstruction and Applications} The aligned depth map $D_{aligned}$ produced by the proposed method provides consistent metric-scale depth maps with significantly improved accuracy on glass surfaces, while preserving the structural details of the monocular depth prior. 
As such, it can be readily integrated into existing RGB-D mapping pipelines for downstream tasks. 
For example, the corrected depth can be consumed by RGB-D SLAM and online dense-mapping systems, such as ORB-SLAM3~\cite{campos2021orbslam3} and VG-Mapping~\cite{he2025vgmapping}, as well as dense reconstruction frameworks such as FrozenRecon~\cite{xu2023frozenrecon}.
These improved reconstructions can then benefit a variety of robotics applications, including semantic mapping, obstacle avoidance, and safe navigation in settings where glass surfaces would otherwise be misperceived. Furthermore, we provide an example of robot navigation in Sec.~\ref{sec:nav} to demonstrate the effectiveness and importance of our proposed method for robotic applications.

\textbf{Modularity and Generality}. 
The proposed alignment module is entirely decoupled from the specific choice of the depth prior, $D_{prior}$, allowing it to serve as a versatile front-end for downstream tasks that rely on metric depth, such as dense SLAM, 3D reconstruction, and visual navigation. Furthermore, the framework is inherently agnostic to the underlying DFMs. As more advanced affine-invariant networks are emerging, they can be seamlessly integrated as \textit{plug-and-play} components without requiring any algorithmic modifications.

\subsection{Affine Depth Alignment via Local RANSAC}
\label{sec:ransac}
The alignment problem can be defined as computing the optimal scale $s$ and shift $t$ that minimize the discrepancy between the $D_{prior} = \{d_i\}$ and the raw sensor depth $D_{raw} = \{d^*_i\}$ as described in~\cite{ranftl2020towards}.
\begin{equation}
    (s,t) = \arg\min_{s,t} \sum_{i} (sd_i + t- d_{i}^{*})^2
\label{equ:lsm}
\end{equation}
However, in this paper, we focus on challenging scenarios in which $D_{raw}$ contains outliers caused by glass surfaces. Directly optimizing using Equation~\ref{equ:lsm} causes failure as demonstrated in Table~\ref{tab:exp1}, where the global represents using all the $D_{raw}$ for optimization.
To overcome this challenge, we adopt a RANSAC‑inspired strategy that samples candidate transformations from local image patches and selects the one with the smallest error over the entire image to obtain a robust alignment estimation.
The rationale is that reliable non‑glass regions dominate the total pixel count, so the globally optimal transformation will naturally fit those regions well, while errors on glass are suppressed by the RANSAC process as depicted in Fig.~\ref{fig:method}.

\textbf{Patch‑wise Sampling}. The image is divided into a $N*N$ grid of patches $\{\mathbf{p}_{ij}\}$. For each patch, we perform $K$ iterations of the following steps:
\subsubsection{Random Pixel Selection} Randomly select $M$ pixel locations within the current patch without any filtering for potential glass pixels, since they are mitigated by the subsequent global validation step.
\subsubsection{Parameter Estimation} Using the $M$ selected points, compute the least‑squares solution for $s$ and $t$ by minimizing Equation~\ref{equ:lsm}.

\subsubsection{Error Computation} Candidate parameters $\{s_{ij}, t_{ij}\}$ are applied to the prior depth $\hat{D_{ij}} = s_{ij}D_{prior} + t_{ij}$. 
Total absolute error is computed over the entire image.

\begin{equation}
\label{equ:error}
    e_{ij} = \sum\limits \lvert \hat{D_{ij}} - D_{raw}\rvert
\end{equation}
\subsubsection{Keep Best} Retain the parameter pair ${s^*, t^*}$ that yields the minimal $e_{ij}$ across all iterations and all patches. After processing all patches, the globally best transformation is applied to obtain metric-aligned depth:
\begin{equation}
\label{equ:align}
    \hat{D^*} = s^* D_{prior} + t^*
\end{equation}

Our patch‑based sampling strategy aims to circumvent the detrimental influence of erroneous depth measurements on glass surfaces. In conventional global alignment, all pixels—including those on glass where the sensor depth is either invalid or corresponds to background objects—contribute to the parameter estimation, leading to biased scale and shift. Our approach mitigates this issue by performing random sampling within local patches. Since glass regions typically occupy a small fraction of the image, the randomly selected pixels within a patch are likely to come from non‑glass areas, where the sensor depth is reliable. The candidate parameters computed from such samples naturally fit the metric depth in reliable regions. By evaluating each candidate's error over the entire image and selecting the one that minimizes this global error, we effectively identify the transformation that best aligns the prior with the correct metric scale, while the contributions from erroneous glass pixels are implicitly discarded.
\begin{figure*}[!h]
\centering
\includegraphics[width=0.95\textwidth]{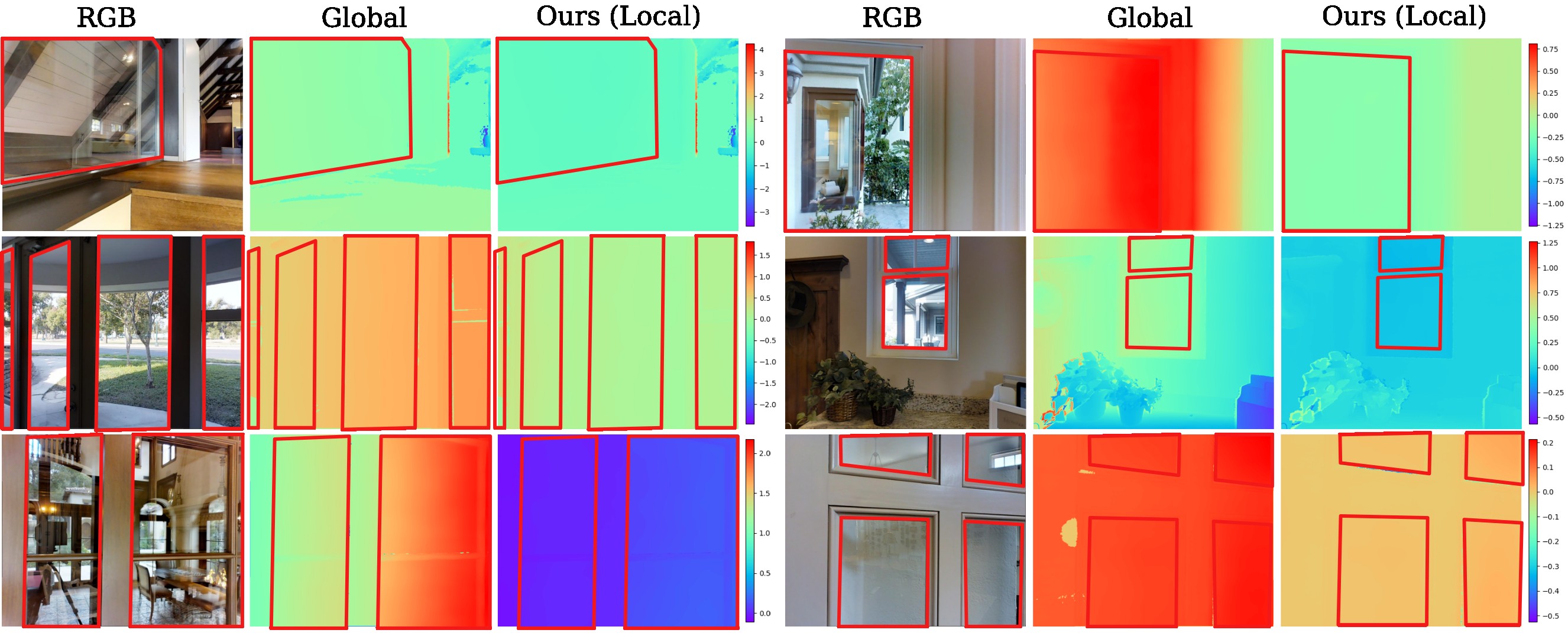}
\caption{\bt{Visualization of error maps for depth alignment results.} Selective samples from \ti{hard} subset of \ti{\bt{GlassRecon}}.} 
\label{fig:exp1}
\end{figure*}
\section{Experimental Results}
\label{sec:exp}
We conduct two sets of experiments to evaluate the effectiveness of our proposed alignment method. 
\begin{itemize}
    \item Sec.~\ref{sec:affine-invariant}: Comparison of different alignment strategies, global alignment, and the proposed patch-wise approach with different affine-invariant depth prediction models.  
    \item Sec.~\ref{sec:metric_depth}: Comparison of the proposed method with SotA metric depth prediction networks.
\end{itemize}
In this section, we first introduce the dataset and evaluation metrics, followed by a detailed experimental analysis.

\subsection{Dataset and Metrics}
We evaluate depth accuracy using two metrics computed over the entire image: Absolute Relative Error (AbsRel) and the accuracy at $\delta < 1.25$ as in~\cite{yang2024depth}.
To evaluate performance across varying difficulties, we partition the 917 images in \ti{\bt{GlassRecon}} into \ti{easy} (601 samples, $\mathrm{AbsRel} \leq 0.03$) and \ti{hard} (316 samples, $\mathrm{AbsRel} > 0.03$) subsets based on the Absolute Relative Error (AbsRel) between $D_{raw}$ and $D_{GT}$.
The \ti{hard} subset consists of samples featuring larger glass regions that exhibit significant errors.

\begin{figure*}[!h]
\centering
\includegraphics[width=0.95\textwidth]{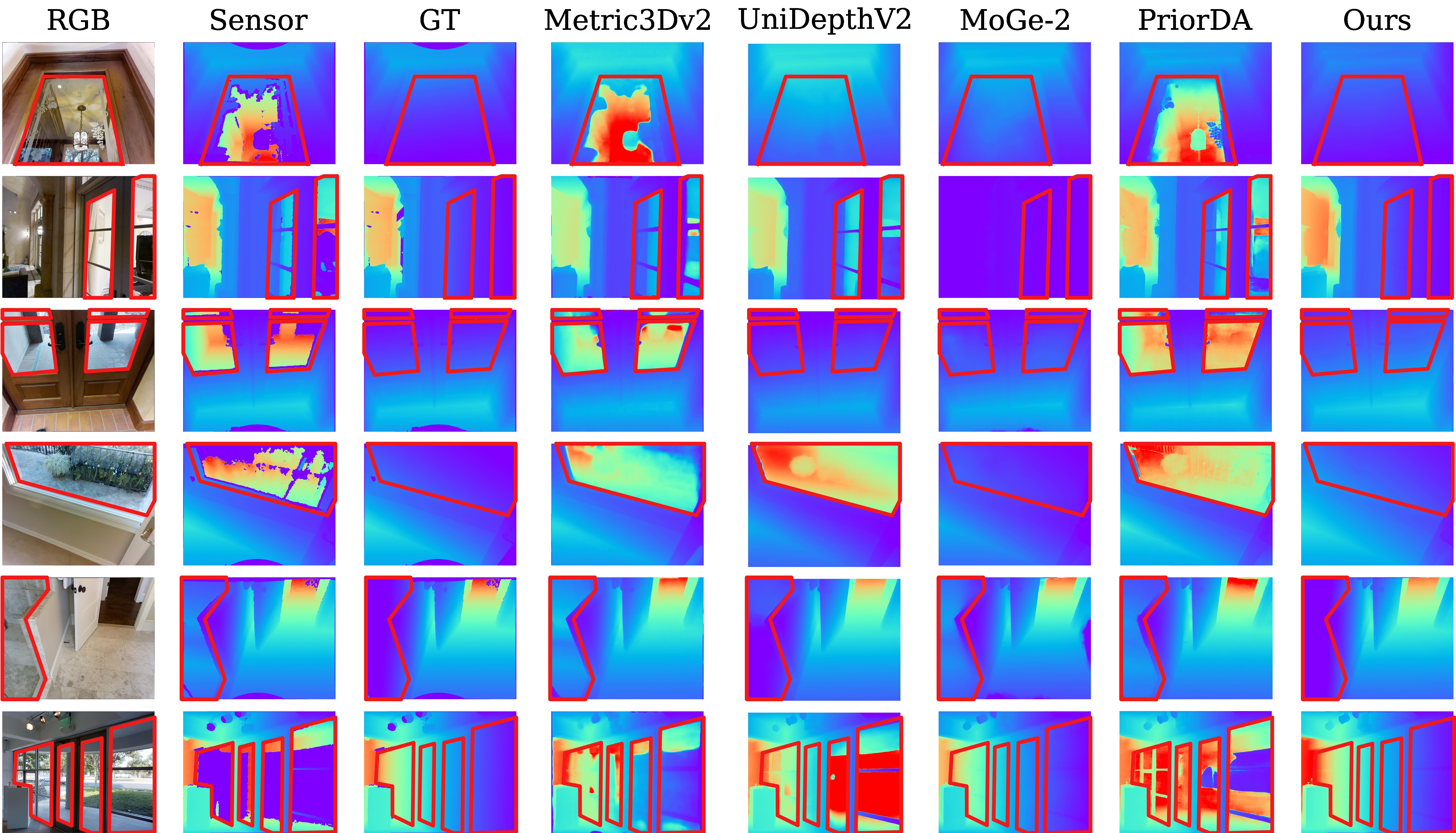}
\caption{\textbf{Comparison with metric depth prediction models.} 
As shown in Table~\ref{tab:exp2}, Metric3Dv2 and UniDepthV2 are demonstrated without intrinsic inputs, whereas PriorDA with the geometry prior.}
\label{fig:exp2}
\end{figure*}
\begin{table}[h!]
\centering
\small
\caption{\textbf{Depth Alignment Results with Relative Depth Estimators}. 
}
\resizebox{0.48\textwidth}{!}{
\begin{tabular}{llcccccc}
\toprule
\multicolumn{2}{c}{\multirow{2}{*}{\textbf{Method}}} & \multicolumn{2}{c}{\textbf{All}} & \multicolumn{2}{c}{\textbf{Easy}} & \multicolumn{2}{c}{\textbf{Hard}} \\ \cmidrule(lr){3-4} \cmidrule(lr){5-6} \cmidrule(lr){7-8}
\multicolumn{2}{c}{} & AbsRel $\downarrow$ & $\delta_1$ $\uparrow$ & AbsRel $\downarrow$ & $\delta_1$ $\uparrow$ & AbsRel $\downarrow$ & $\delta_1$ $\uparrow$ \\ \midrule

\multirow{2}{*}{DAV2~\cite{yang2024depth}} & Global & 0.167 & 0.876 & 0.112 & \textbf{0.937} & 0.272 & 0.762 \\
 & Ours (Local) & \textbf{0.158} & \textbf{0.906} & \textbf{0.100} & 0.936 & \textbf{0.269} & \textbf{0.849} \\ \midrule

\multirow{2}{*}{MoGe~\cite{wang2025moge}} & Global & 0.166 & 0.874 & 0.059 & 0.963 & 0.369 & 0.703 \\
& Ours (Local) &\textbf{0.142} & \textbf{0.904} & \textbf{0.055} & \textbf{0.964} & \textbf{0.309} & \textbf{0.790} \\ \midrule
 
\multirow{2}{*}{DA3~\cite{depthanything3}} & Global & 0.152 & 0.855 & 0.061 & 0.962 & 0.323 & 0.651 \\
& Ours (Local) & \textbf{0.095} & \textbf{0.937} & \textbf{0.055} & \textbf{0.966} & \textbf{0.172} & \textbf{0.883} \\
\bottomrule
\label{tab:exp1}
\end{tabular}
}
\end{table}
\subsection{Comparison with Affine-Invariant Depth Methods}
\label{sec:affine-invariant}
We first evaluate the effectiveness of our alignment strategy on three SotA affine-invariant depth prediction networks: DAV2~\cite{yang2024depth}, DA3~\cite{depthanything3}, and MoGe~\cite{wang2025moge}. 
As shown in Table~\ref{tab:exp1}, our proposed local RANSAC-alignment method consistently outperforms the global alignment baseline across almost all networks and subsets, with the most substantial gains observed on the hard subset where glass causes severe depth corruption. 
Among three networks, DA3 achieves the best overall performance when combined with our method, reducing AbsRel on the hard subset by over 46\%.
The qualitative results in Fig.~\ref{fig:exp1} illustrate the depth difference between our method and global alignment against ground truth for representative samples from the hard subset. 
In these examples, the raw sensor depth incorrectly assigns background values to the glass surface, leading to a heavily biased global alignment and large errors throughout the glass region. On the other hand, our local RANSAC alignment successfully recovers the correct planar structure, producing an error map that is substantially lower than that of the global alignment, consistent with the quantitative gains reported in Table~\ref{tab:exp1} for \ti{hard} subset.
\begin{table}[h!]

\centering
\normalsize
\caption{\textbf{Metric Depth Estimation Result.} 
\firsttxt{\textbf{Best}}, \secondtxt{\textbf{Second Best}}, and \thirdtxt{\textbf{Third Best}} are highlighted. 
}
\resizebox{0.48\textwidth}{!}{
\begin{tabular}{llcccccc}
\toprule
\multicolumn{2}{c}{\multirow{2}{*}{\textbf{Method}}} & \multicolumn{2}{c}{\textbf{All}} & \multicolumn{2}{c}{\textbf{Easy}} & \multicolumn{2}{c}{\textbf{Hard}} \\ \cmidrule(lr){3-4} \cmidrule(lr){5-6} \cmidrule(lr){7-8}
\multicolumn{2}{c}{} & AbsRel $\downarrow$ & $\delta_1$ $\uparrow$ & AbsRel $\downarrow$ & $\delta_1$ $\uparrow$ & AbsRel $\downarrow$ & $\delta_1$ $\uparrow$ \\ \midrule

\multirow{2}{*}{Metric3Dv2~\cite{hu2024metric3d}}  & w/o intrin. & 0.245 & 0.834 & 0.168 & 0.897 & 0.390 & 0.713 \\
  & w/ intrin. & 0.258 & 0.819 & \thirdc \textbf{0.167} & 0.887 & 0.432 & 0.690 \\ \midrule

\multirow{2}{*}{UniDepthV2~\cite{piccinelli2025unidepthv2}} & w/o intrin. & 0.305 & 0.804 & 0.204 & 0.865 & 0.496 & 0.687 \\
  & w/ intrin. & 0.309 & 0.808 & 0.204 & 0.872 & 0.509 & 0.686 \\ \midrule

MoGe-2~\cite{wang2025moge2}  & & 0.240 & 0.540 & 0.229 & 0.548 & \secondc \textbf{0.262} & 0.524 \\ \midrule

\multirow{2}{*}{PriorDA~\cite{wang2025depth}}  & w/o geom. & \thirdc \textbf{0.158} & \thirdc \textbf{0.886} & \firstc \textbf{0.037} & \secondc \textbf{0.970} & 0.387 & \secondc \textbf{0.726} \\
 & w/ geom. & \secondc \textbf{0.155} & \secondc \textbf{0.887} & \firstc \textbf{0.037} &\firstc \textbf{0.972} & \thirdc \textbf{0.381} & \thirdc \textbf{0.724} \\ \midrule

\multicolumn{2}{c}{Ours (+DA3~\cite{depthanything3})}& \firstc \textbf{0.095} & \firstc \textbf{0.937} & \secondc \textbf{0.055} & \thirdc \textbf{0.966} & \firstc \textbf{0.172} & \firstc \textbf{0.883} \\ \bottomrule
\label{tab:exp2}
\end{tabular}
}
\end{table}
\subsection{Comparison with Metric Depth Methods}
\label{sec:metric_depth}
Based on the benchmark results and analysis in Sec.~\ref{sec:affine-invariant}, we choose DA3 as the depth prior in combination with our method to compare with metric depth prediction models.
As summarized in Table~\ref{tab:exp2}, our method achieves the best overall performance, with significant improvement on the \ti{hard} split where glass-induced errors are severe. 
On the \ti{easy} subset, our method achieves competitive performance with PriorDA, which is designed to refine raw sensor depth without accounting for potential measurement corruption, as shown in Fig.~\ref{fig:exp2}.
Furthermore, from the visualization of selective samples in \ti{hard} subset, we find that PriorDA remains anchored to the input sensor depth, failing to recover the true glass geometry. While Metric3Dv2 and UniDepthV2 similarly struggle to estimate correct structures on glass, MoGe-2 achieves more plausible geometry but suffers from frequent depth scale misalignment. In contrast, our method successfully recovers a clean, planar glass surface with an accurate metric scale that closely matches the ground truth. This visual accuracy is entirely consistent with our quantitative advantage on the \ti{hard} subset, as demonstrated in Table~\ref{tab:exp2}. 

\begin{figure}[ht]
\centering
\includegraphics[width=0.48\textwidth]{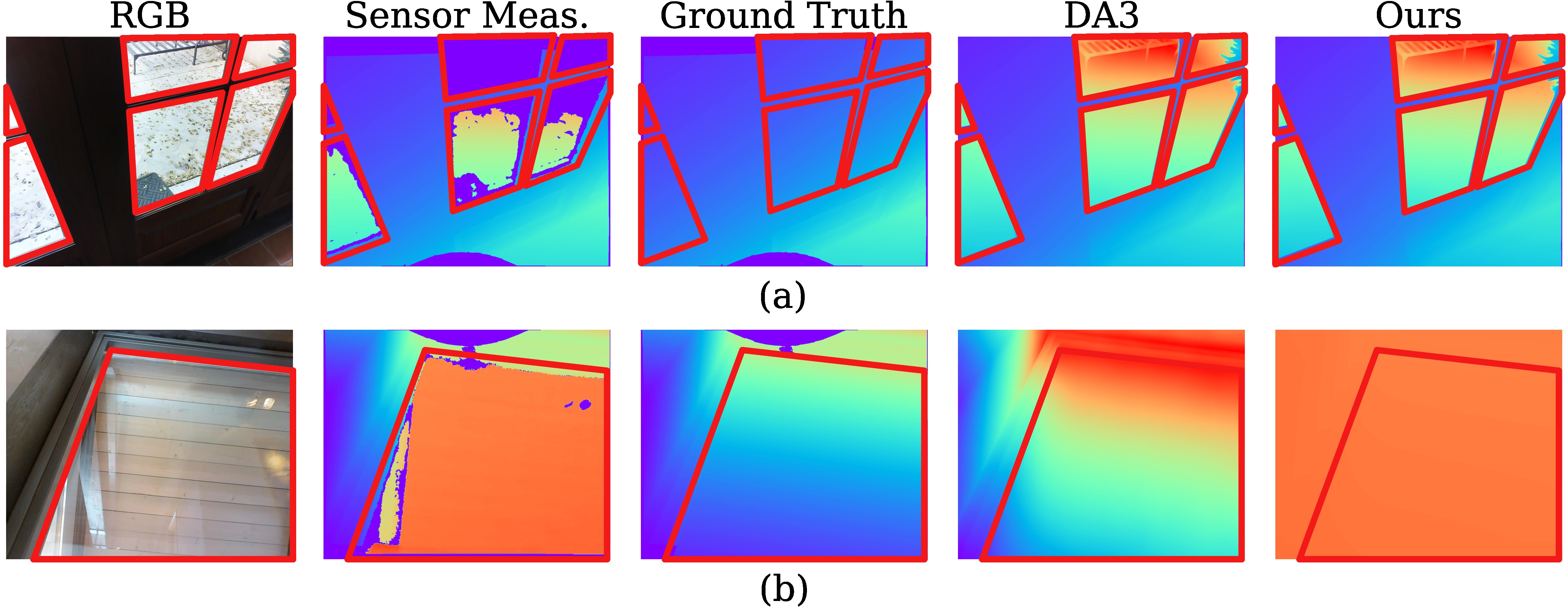}
\caption{\textbf{Visualization of representative failure cases.} (a) 
Limitations of the monocular depth prior.
(b) Corruption from erroneous sensor depth.
}
\label{fig:failure}
\end{figure}
\subsection{Limitations and Failure Cases}
Despite the effectiveness of our method on the majority of glass surfaces, two primary scenarios still pose challenges as illustrated in Fig.~\ref{fig:failure}.

\textbf{Limitations of the depth prior}. Our approach scales the output of the depth prediction network to improve the reconstruction of glass surfaces.
If the depth prior is limited to predict the geometry of glass regions, our proposed method fails as shown in Fig.~\ref{fig:failure}(a), where the prior incorrectly estimates depth corresponding to background objects behind the glass rather than the glass surface.
Though this highlights the dependence of our method on the quality of the underlying depth prior, the emerging developments in DFMs would seamlessly benefit our proposed framework.

\textbf{Corruption from erroneous sensor depth}. The robustness of our local RANSAC alignment stems from the assumption that randomly sampled pixels are mainly from regions where the sensor depth is reliable. This assumption holds when glass regions either occupy a small area or return mostly invalid measurements (e.g., zeros or missing values). However, when depth sensor measurements return erroneous but valid depth values over a glass surface that occupies most of the image, these corrupted pixels can dominate the global error minimization process. 
The resulting scale and shift are then biased toward fitting the incorrect measurements, causing the aligned depth to conform to the erroneous values rather than the true glass geometry. 
Fig.~\ref{fig:failure}~(b) illustrates a typical case, where large glass areas with incorrect depth values bias the alignment. One potential improvement could be to modify the global error minimization to better handle dominant erroneous depths. For example, by introducing per-pixel weighting or modifying the validation strategy. We leave this direction for future work.
\begin{figure*}[ht]
\centering
\includegraphics[width=0.95\textwidth]{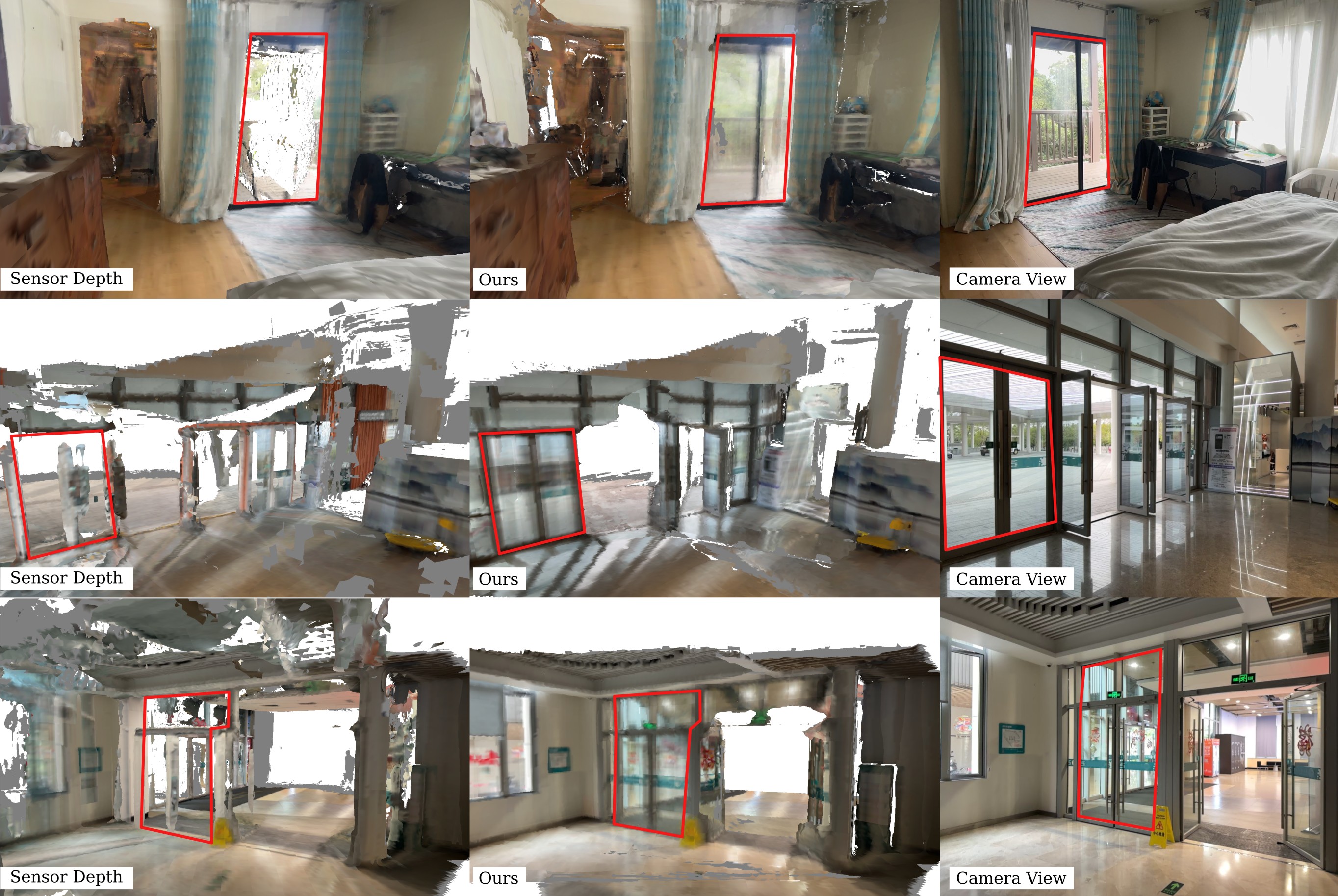}
\caption{\textbf{Visualization of Reconstruction.} (Top) ScanNet++, (Middle) Scene01, (bottom) Scene02. The top row is selected from ScanNet++~\cite{yeshwanth2023scannet++}, and the Scene01 and Scene02 are from self-collected data sequences with an iPhone. 
}
\label{fig:recon}
\end{figure*}
\section{Application in Robot Navigation}
\label{sec:nav}

To demonstrate the practical effectiveness of our method for downstream robotic applications, we evaluate its impact on mapping and navigation through two complementary experiments: one utilizing the public ScanNet++~\cite{yeshwanth2023scannet++} dataset, and the other using real-world data collected in campus environments.

\textbf{Evaluation on ScanNet++.} We first assess 3D reconstruction quality on a challenging indoor scene from ScanNet++~\cite{yeshwanth2023scannet++} featuring a glass balcony door, a common failure point for standard depth sensors. Using the nvblox~\cite{millane2024nvblox} reconstruction framework, we compare the map generated from raw sensor depth against the map produced with our corrected depth. As shown in the top row of Fig.~\ref{fig:recon}, the reconstruction relying on raw sensor depth exhibits severe holes around the glass door, rendering the area unsafe for navigation. In contrast, our corrected depth successfully recovers the planar structure of the glass surface, yielding a complete and geometrically consistent map.

\begin{figure}[h]
\centering
\includegraphics[width=0.48\textwidth]{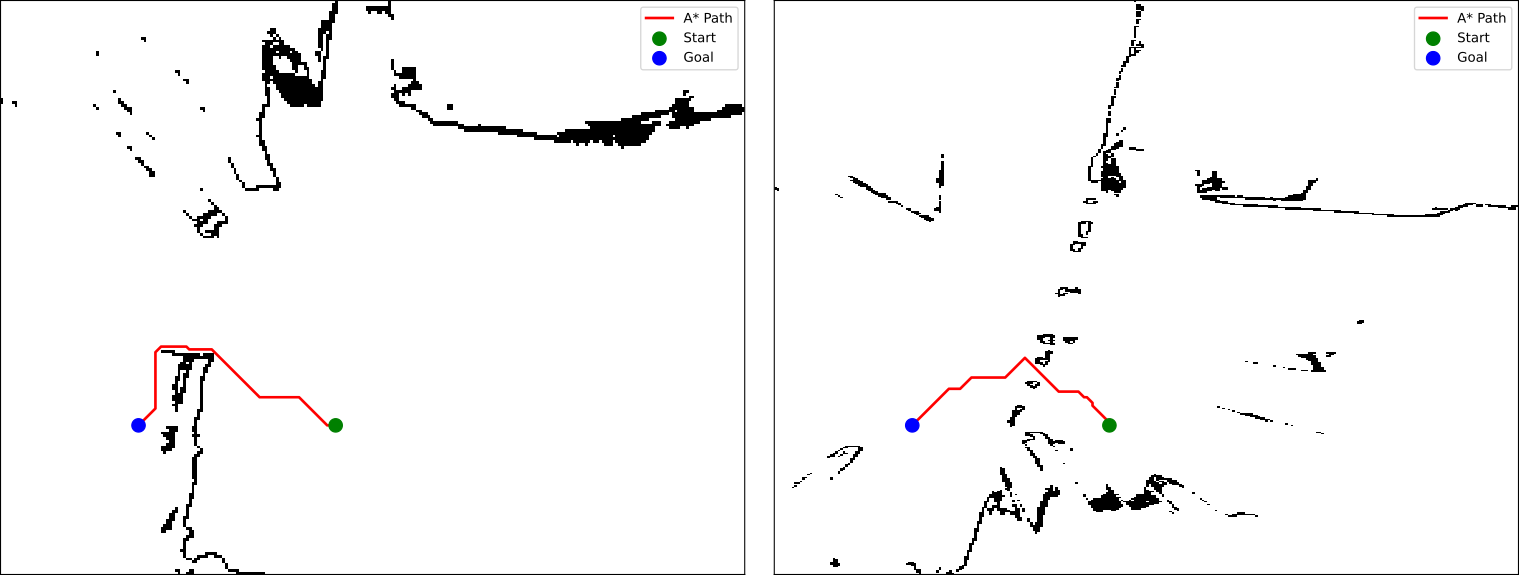}
\caption{\textbf{Visualization of the planning result on Scene02.} Left: Motion planning result on our reconstructed map. Right: Motion planning result on the sensor depth reconstructed map. 
}
\label{fig:planning}
\end{figure}
\textbf{Real-World Validation.} To validate our approach in the wild, we conducted field experiments in the lobbies of two campus buildings. Using an iPhone 17 Pro, we recorded two RGB-D sequences (with associated camera poses) in areas featuring multiple glass doors. After applying our depth correction to each downsampled frame, we reconstructed 3D maps of both environments. The middle and bottom rows of Fig.~\ref{fig:recon} illustrate the reconstruction results for Scene01 and Scene02, respectively. Consistent with our ScanNet++ findings, raw sensor depth treats glass regions as empty voids, whereas our corrected depth reconstructs these areas with complete, accurate geometry.

Map fidelity is especially important for planning methods whose candidate edges are validated through collision checking~\cite{huang2026lifelong}. To further demonstrate the impact on robot navigation, we converted the reconstructed point cloud of Scene02 (Fig.~\ref{fig:recon}, bottom row) into a 2D occupancy grid for path planning. Fig.~\ref{fig:planning} displays the planned trajectories overlaid on this map. When using the map generated from raw sensor depth, the planner misinterprets the glass doors as traversable open space, resulting in a collision-prone path. Conversely, our corrected depth accurately registers the glass surfaces as physical obstacles, enabling the planner to compute a safe, collision-free route. 

In summary, these experiments confirm that our depth correction not only improves quantitative depth metrics but also provides tangible benefits for real-world robotics, enabling robust scene reconstruction and safe path planning in glass-rich environments.

\section{Conclusions and Future Work}
In this paper, we present a simple yet effective method to achieve glass surface depth completion that fuses affine-invariant depth priors
with raw sensor measurement. 
By applying a RANSAC-based sampling strategy, our approach naturally avoids contamination from erroneous glass depths while preserving accurate geometry without training. 
To support evaluation, we build \ti{\bt{GlassRecon}}, which consists of 917 indoor images with ground-truth depth annotations, including the glass regions. 
Through extensive experiments, we find that leveraging DA3 as prior in the proposed framework outperforms SotA metric depth prediction models, with particularly pronounced gains on \ti{hard} examples where sensor depth is severely corrupted.

For future work, we plan to expand the proposed dataset to encompass a wider variety of glass types and scenes. This will not only establish it as a comprehensive benchmark for glass surface depth estimation but also enable the fine-tuning of depth priors for enhanced performance. Furthermore, incorporating uncertainty estimation into the alignment process could improve robustness by adaptively weighting pixels based on their reliability. 
Finally, inspired by trajectory priors that have proven effective in rejecting inconsistent SLAM constraints~\cite{yu2026rover}, we plan to exploit temporal consistency across sequential RGB-D frames to identify unreliable per-frame depth alignments and resolve geometric ambiguities when single-frame priors are insufficient.

\bibliographystyle{IEEEtran}
\bibliography{ref}

@STRING{ICRA = "Proc. {IEEE} Int. Conf. Robot. Autom."}

@STRING{ICMA ="Proc. {IEEE} Int. Conf. Mechatronics Autom." }

@STRING{CVPR = "Proc. {IEEE} Conf. Comput. Vis. Pattern Recognit."}

@STRING{IJRR = "Int. J. Robot. Res. (IJRR)"}

@STRING{PAMI = "{IEEE} Trans. Pattern Anal. Mach. Intell."}

@STRING{TRO = "{IEEE} Trans. Robot."}

@STRING{TASE = "{IEEE} Autom. Sci. Eng."}

@STRING{TIP = "{IEEE} Trans. Image Process."}

@STRING{NIPS = "Adv. Neural Inf. Process. Syst."}

@STRING{ICCV = "Proc. {IEEE} Int. Conf. Comput. Vis."}

@STRING{RAS = "Robot. and Auton. Syst."}

@STRING{TIM = "{IEEE} Trans. Instrum. Meas."}

@STRING{DV3 = "Int. Conf. 3D Vis."}

@STRING{AAAI = "Int. Conf. Learning Representations"}

@article{campos2021orbslam3,
  title   = {{ORB-SLAM3}: An Accurate Open-Source Library for Visual, Visual-Inertial, and Multimap {SLAM}},
  author  = {Campos, Carlos and Elvira, Richard and Rodr{\'i}guez, Juan J. G{\'o}mez and Montiel, Jos{\'e} M. M. and Tard{\'o}s, Juan D.},
  journal = TRO,
  volume  = {37},
  number  = {6},
  pages   = {1874--1890},
  year    = {2021},
  doi     = {10.1109/TRO.2021.3075644}
}

@article{he2025vgmapping,
  title   = {{VG-Mapping}: Variation-Aware Density Control for Online 3D Gaussian Mapping in Semi-Static Scenes},
  author  = {He, Yicheng and Yu, Jingwen and Chen, Guangcheng and Zhang, Hong},
  journal = {arXiv preprint arXiv:2510.09962},
  year    = {2025},
  doi     = {10.48550/arXiv.2510.09962}
}

@article{huang2026lifelong,
  title   = {Asymptotically Optimal Lifelong Planning With Lazy Edge Evaluation Under Expensive Collision Checks},
  author  = {Huang, Lu and Yu, Jingwen and Wang, Jiankun and Jing, Xingjian},
  journal = TASE,
  volume  = {23},
  pages   = {401--416},
  year    = {2026},
  doi     = {10.1109/TASE.2025.3636173}
}

@article{yu2026rover,
  title   = {{ROVER}: Robust Loop Closure Verification With Trajectory Prior in Repetitive Environments},
  author  = {Yu, Jingwen and Yang, Jiayi and Jiao, Jianhao and Hu, Anjun and Liu, Zhonghang and Wang, Jiankun and Tan, Ping and Zhang, Hong},
  journal = TASE,
  volume  = {23},
  pages   = {8476--8489},
  year    = {2026},
  doi     = {10.1109/TASE.2026.3684100}
}

@article{wei2025fusionportablev2,
  title={FusionPortableV2: A unified multi-sensor dataset for generalized SLAM across diverse platforms and scalable environments},
  author={Wei, Hexiang and Jiao, Jianhao and Hu, Xiangcheng and Yu, Jingwen and Xie, Xupeng and Wu, Jin and Zhu, Yilong and Liu, Yuxuan and Wang, Lujia and Liu, Ming},
  journal= IJRR,
  volume={44},
  number={7},
  pages={1093--1116},
  year={2025},
  publisher={Sage Publications Sage UK: London, England}
}

@article{ranftl2020towards,
  title={Towards robust monocular depth estimation: Mixing datasets for zero-shot cross-dataset transfer},
  author={Ranftl, Ren{\'e} and Lasinger, Katrin and Hafner, David and Schindler, Konrad and Koltun, Vladlen},
  journal=PAMI,
  volume={44},
  number={3},
  pages={1623--1637},
  year={2020},
  publisher={IEEE}
}

@inproceedings{lin2025leveraging,
  title={Leveraging RGB-D data with cross-modal context mining for glass surface detection},
  author={Lin, Jiaying and Yeung, Yuen-Hei and Ye, Shuquan and Lau, Rynson WH},
  booktitle={Proceedings of the AAAI Conference on Artificial Intelligence},
  volume={39},
  number={5},
  pages={5254--5261},
  year={2025}
}

@article{piccinelli2025unidepthv2,
  title={Unidepthv2: Universal monocular metric depth estimation made simpler},
  author={Piccinelli, Luigi and Sakaridis, Christos and Yang, Yung-Hsu and Segu, Mattia and Li, Siyuan and Abbeloos, Wim and Van Gool, Luc},
  journal={arXiv preprint arXiv:2502.20110},
  year={2025}
}

@inproceedings{wang2025moge,
  title={Moge: Unlocking accurate monocular geometry estimation for open-domain images with optimal training supervision},
  author={Wang, Ruicheng and Xu, Sicheng and Dai, Cassie and Xiang, Jianfeng and Deng, Yu and Tong, Xin and Yang, Jiaolong},
  booktitle=CVPR,
  pages={5261--5271},
  year={2025}
}

@article{yang2024depth,
  title={Depth anything v2},
  author={Yang, Lihe and Kang, Bingyi and Huang, Zilong and Zhao, Zhen and Xu, Xiaogang and Feng, Jiashi and Zhao, Hengshuang},
  journal=NIPS,
  volume={37},
  pages={21875--21911},
  year={2024}
}

@article{hu2024metric3d,
  title={Metric3d v2: A versatile monocular geometric foundation model for zero-shot metric depth and surface normal estimation},
  author={Hu, Mu and Yin, Wei and Zhang, Chi and Cai, Zhipeng and Long, Xiaoxiao and Chen, Hao and Wang, Kaixuan and Yu, Gang and Shen, Chunhua and Shen, Shaojie},
  journal=PAMI,
  year={2024},
  publisher={IEEE}
}

@article{Matterport3D,
  title={Matterport3D: Learning from RGB-D Data in Indoor Environments},
  author={Chang, Angel and Dai, Angela and Funkhouser, Thomas and Halber, Maciej and Niessner, Matthias and Savva, Manolis and Song, Shuran and Zeng, Andy and Zhang, Yinda},
  journal=DV3,
  year={2017}
}

@article{wang2025moge2,
  title={MoGe-2: Accurate Monocular Geometry with Metric Scale and Sharp Details},
  author={Wang, Ruicheng and Xu, Sicheng and Dong, Yue and Deng, Yu and Xiang, Jianfeng and Lv, Zelong and Sun, Guangzhong and Tong, Xin and Yang, Jiaolong},
  journal={arXiv preprint arXiv:2507.02546},
  year={2025}
}

@article{wang2025depth,
  title={Depth Anything with Any Prior},
  author={Wang, Zehan and Chen, Siyu and Yang, Lihe and Wang, Jialei and Zhang, Ziang and Zhao, Hengshuang and Zhao, Zhou},
  journal={arXiv preprint arXiv:2505.10565},
  year={2025}
}

@article{depthanything3,
  title={Depth Anything 3: Recovering the visual space from any views},
  author={Haotong Lin and Sili Chen and Jun Hao Liew and Donny Y. Chen and Zhenyu Li and Guang Shi and Jiashi Feng and Bingyi Kang},
  journal={arXiv preprint arXiv:2511.10647},
  year={2025}
}

@inproceedings{xu2023frozenrecon,
  title={Frozenrecon: Pose-free 3d scene reconstruction with frozen depth models},
  author={Xu, Guangkai and Yin, Wei and Chen, Hao and Shen, Chunhua and Cheng, Kai and Zhao, Feng},
  booktitle=ICCV,
  pages={9276--9286},
  year={2023},
  organization={IEEE}
}

@article{wang2017detecting,
  title = {Detecting glass in simultaneous localisation and mapping},
  author = {Wang, X. and Wang, J.},
  journal = RAS,
  volume = {88},
  pages = {97--103},
  year = {2017},
  key = {wang2017}
}

@article{zhou2024lidar,
  title={LiDAR-based 3-D glass detection and reconstruction in indoor environment},
  author={Zhou, Lelai and Sun, Xiaohui and Zhang, Chen and Cao, Luyang and Li, Yibin},
  journal=TIM,
  volume={73},
  pages={1--11},
  year={2024},
  publisher={IEEE}
}

@article{zhang2017three,
  title = {3D reconstruction in the presence of glass and mirrors by acoustic and visual fusion},
  author = {Zhang, Y. and Ye, M. and Manocha, D. and Yang, R.},
  journal = PAMI,
  volume = {40},
  number = {8},
  pages = {1785--1798},
  year = {2017},
  key = {zhang2017three}
}

@inproceedings{mei2022glass,
  title = {Glass segmentation using intensity and spectral polarization cues},
  author = {Mei, H. and Dong, B. and Dong, W. and Yang, J. and Baek, S. H. and Heide, F. and ... and Yang, X.},
  year = {2022},
  booktitle = CVPR,
  pages = {12622-12631},
  note = {DOI: }
}

@article{huo2023glass,
  title = {Glass segmentation with RGB-thermal image pairs},
  author = {Huo, D. and Wang, J. and Qian, Y. and Yang, Y. H.},
  journal = TIP,
  volume = {32},
  pages = {1911--1926},
  year = {2023},
  doi = {},
  issn = {1057-7149},
  language = {English}
}

@inproceedings{zhu2021transfusion,
  title = {Transfusion: A novel SLAM method focused on transparent objects},
  author = {Zhu, Y. and Qiu, J. and Ren, B.},
  year = {2021},
  booktitle = CVPR,
  pages = {6019--6028},
  note = {key={zhu2021}}
}

@inproceedings{zhao2023glass,
  title = {Glass detection in simultaneous localization and mapping of mobile robot based on RGB-D camera},
  author = {Zhao, Y. and Li, H. and Jiang, S. and Li, H. and Zhang, Z. and Zhu, H.},
  year = {2023},
  month = {aug},
  booktitle = ICMA,
  pages = {549-556},
  publisher = {IEEE},
  note = {DOI: },
  key = {zhao2023}
}

@article{liang2023monocular,
  title={Monocular depth estimation for glass walls with context: a new dataset and method},
  author={Liang, Yuan and Deng, Bailin and Liu, Wenxi and Qin, Jing and He, Shengfeng},
  journal=PAMI,
  volume={45},
  number={12},
  pages={15081--15097},
  year={2023},
  publisher={IEEE}
}

@article{zhang2025monoglass3d,
  title={MonoGlass3D: Monocular 3D Glass Detection with Plane Regression and Adaptive Feature Fusion},
  author={Zhang, Kai and Zhao, Guoyang and Shi, Jianxing and Liu, Bonan and Qi, Weiqing and Ma, Jun},
  journal={arXiv preprint arXiv:2509.05599},
  year={2025}
}

@inproceedings{yang2024depth-1,
  title = {Depth anything: Unleashing the power of large-scale unlabeled data},
  author = {Yang, L. and Kang, B. and Huang, Z. and Xu, X. and Feng, J. and Zhao, H.},
  year = {2024},
  booktitle = CVPR,
  pages = {10371-10381},
  note = {DOI: }
}

@inproceedings{yeshwanth2023scannet++,
  title={Scannet++: A high-fidelity dataset of 3d indoor scenes},
  author={Yeshwanth, Chandan and Liu, Yueh-Cheng and Nie{\ss}ner, Matthias and Dai, Angela},
  booktitle=ICCV,
  pages={12--22},
  year={2023}
}

@inproceedings{millane2024nvblox,
  title={nvblox: Gpu-accelerated incremental signed distance field mapping},
  author={Millane, Alexander and Oleynikova, Helen and Wirbel, Emilie and Steiner, Remo and Ramasamy, Vikram and Tingdahl, David and Siegwart, Roland},
  booktitle=ICRA,
  pages={2698--2705},
  year={2024},
  organization={IEEE}
}
\end{document}